\documentclass[lettersize,journal]{IEEEtran}
\usepackage{amsmath,amsfonts}
\usepackage{algorithmic}
\usepackage{algorithm}
\usepackage{array}
\usepackage[caption=false,font=normalsize,labelfont=sf,textfont=sf]{subfig}
\usepackage{textcomp}
\usepackage{stfloats}
\usepackage{url}
\usepackage{verbatim}
\usepackage{graphicx}
\usepackage{cite}
\usepackage{booktabs} 
\usepackage{multirow} 
\usepackage{hyperref} 

\begin{document}

\title{Distilling Textual Priors from LLM to Efficient Image Fusion}

\author{Ran Zhang, Xuanhua He, Ke Cao, Liu Liu, \\
         Li Zhang, Man Zhou, Jie Zhang
\thanks{R. Zhang and L. Liu are with Hefei University of Technology, Hefei, China (e-mail: 2023212219@mail.hfut.edu.cn; liuliu@hfut.edu.cn).}
\thanks{X. He, K. Cao, L. Zhang, and M. Zhou are with the University of Science and Technology of China, Hefei, China (e-mail: hexuanhua@mail.ustc.edu.cn; caoke200820@mail.ustc.edu.cn; zanly20@mail.ustc.edu.cn; manman@mail.ustc.edu.cn).}
\thanks{J. Zhang is with Hefei Institutes of Physical Science, Chinese Academy of Sciences, Hefei, China (e-mail: zhangjie@iim.ac.cn).}}


\maketitle

\begin{abstract}
Multi-modality image fusion aims to synthesize a single, comprehensive image from multiple source inputs. 
Traditional approaches, such as CNNs and GANs, offer efficiency but struggle to handle low-quality or complex inputs. Recent advances in text-guided methods leverage large model priors to overcome these limitations, but at the cost of significant computational overhead, both in memory and inference time. To address this challenge, we propose a novel framework for distilling large model priors, eliminating the need for text guidance during inference while dramatically reducing model size. Our framework utilizes a teacher-student architecture, where the teacher network incorporates large model priors and transfers this knowledge to a smaller student network via a tailored distillation process. Additionally, we introduce a spatial-channel cross-fusion module to enhance the model's ability to leverage textual priors across both spatial and channel dimensions. Our method achieves a favorable trade-off between computational efficiency and fusion quality. The distilled network, requiring only 10\% of the parameters and inference time of the teacher network, retains 90\% of its performance and outperforms existing SOTA methods. Extensive experiments demonstrate the effectiveness of our approach. Codes are available at \url{https://github.com/Zirconium233/DTPF}
\end{abstract}


\section{Introduction}
\IEEEPARstart{I}{mage} fusion plays an important role in visual enhancement within digital image processing. For example, visible-infrared image fusion combines color-based details from visible images, which are easily interpretable by humans, with radiation-based features from infrared images, which are highly effective for target detection and operations under low-light conditions. By integrating complementary information from both modalities, this fusion method produces high-quality images that enhance both human visual interpretation and machine-based detection performance. This work also shares similarities with pan-sharpening, a multi-modal image fusion task that leverages frequency domain analysis, as explored in recent studies such as~\cite{Zhang2025tcsvt}.

\begin{figure}[!t]
    \centering
    \includegraphics[width=\linewidth]{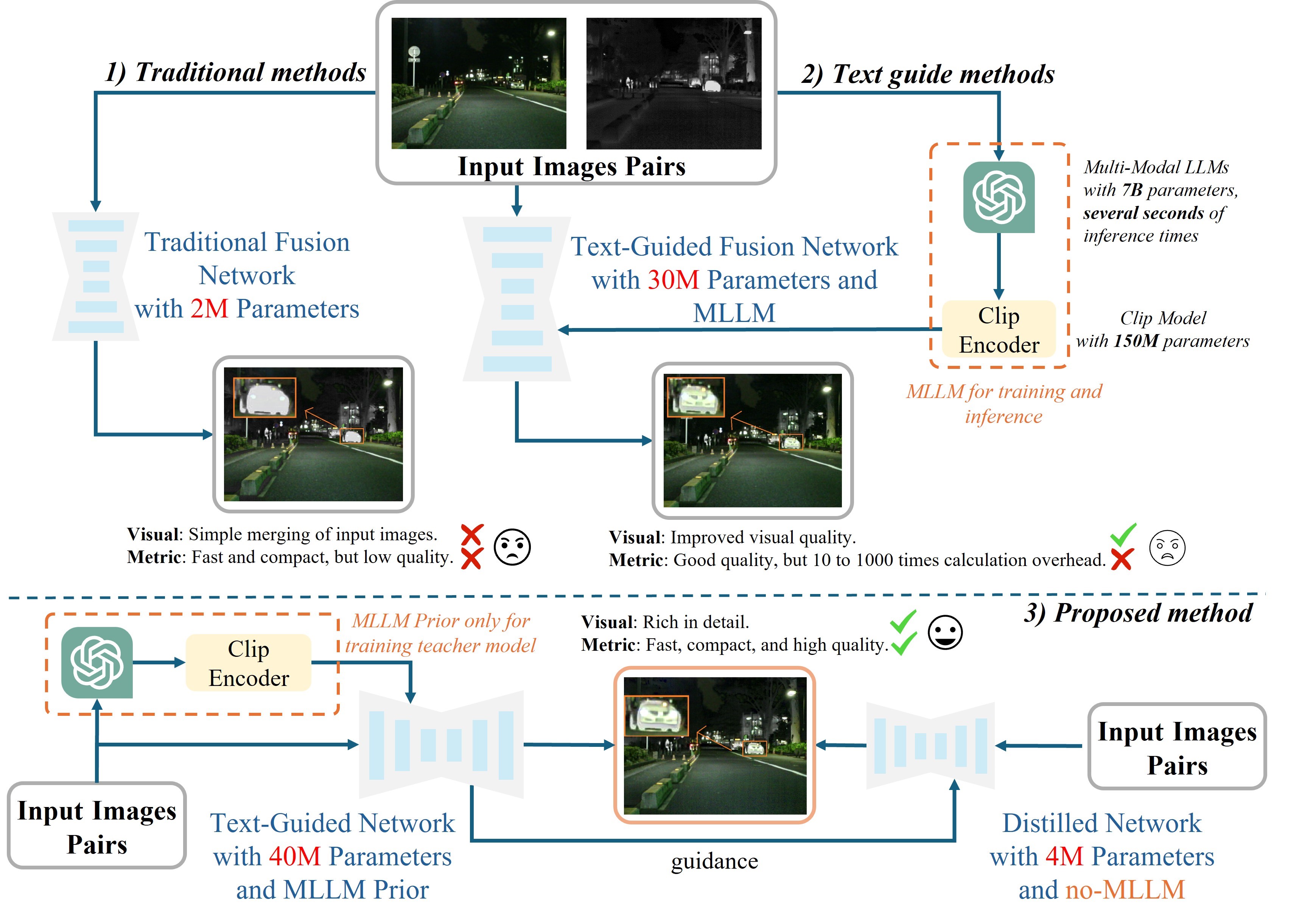}
    \caption{Overview of different image fusion methods and their parameter efficiency. (1) Traditional methods use small fusion networks. (2) Text-guided methods significantly increasing computational demands with performance improvements. (3) Our proposed method leverages text-guided training and knowledge distillation to create a distilled network that achieves high-quality fused images without relying on LLMs during inference.}
    \label{fig:abstract}
\end{figure}
During the imaging process, environmental and device-related constraints often degrade the quality of source images. Visible images may suffer from low resolution, while infrared images are prone to various noise types. Traditional fusion methods, including FusionGAN~\cite{ma2019fusionGAN}, U2Fusion~\cite{xu2020u2fusion}, and SwinFusion~\cite{ma2022swinfusion}, fuse source images without effectively distinguishing between degraded information and meaningful image content, leading to suboptimal performance. Some approaches~\cite{tang2022piafu} mitigate these issues by relying on manual pre-processing to enhance source images, but their flexibility is limited. Recently, text-guided methods, such as TextFusion~\cite{cheng2023textfusion} and Text-IF~\cite{yi2024text}, have emerged as a promising solution to address these challenges. These methods leverage the capabilities of multimodal large language models (MLLMs) to generate text captions or degradation descriptions (e.g., low light) for source images, enabling the model to adaptively distinguish image content from degradation and enhance image quality. By integrating these priors, the model gains enhanced semantic understanding of the image content, which effectively promotes the fusion processes.
For instance, Text-IF utilizes MLLMs~\cite{Qwen-VL} to generate task lists and incorporates CLIP~\cite{clip} to guide the fusion process based on these descriptions. Similarly, TextFusion~\cite{cheng2023textfusion} employs text-based interactions to build models capable of focusing on specific image elements, such as emphasizing trees over people, based on captions generated by LLMs or human input.

Despite their advantages, incorporating LLMs into the fusion workflow introduces significant resource overhead. In most cases, users aim to fuse images for downstream tasks such as detection or segmentation, without the need for extensive interaction. LLMs dramatically increase computational demands, often occupying ten to thousands of times more resources than the fusion module itself. Even CLIP, a relatively lightweight vision-language model, requires approximately five times the parameters of the fusion module. This computational cost makes LLMs disproportionate to the task. Moreover, due to the limitations in the quality and quantity of datasets, the performance gains from integrating large models are marginal compared to the significant decrease in efficiency. Thus, we seek a method that enables the model to perform degradation-aware fusion and enhances semantic understanding without relying on large-scale models.

In response to these issues, we propose a large-model prior distillation framework that eliminates the need for text guidance during inference while significantly reducing model size, as shown in Fig.~\ref{fig:abstract}.
Our framework employs a teacher-student architecture, where the teacher network integrates MLLM priors via feature modulation operations. To fully exploit textual priors across both spatial and channel dimensions, we introduce a spatial-channel cross-fusion module (SCFM). A specially tailored prior distillation loss is then used to transfer textual knowledge from both feature and output perspectives, enabling the student network to mimic the teacher's feature processing without requiring textual inputs. By leveraging this approach, our method achieves a 90\% reduction in model size (excluding CLIP, which alone requires five times the parameters of the original fusion module) while retaining 90\% of the teacher model’s performance.

Our contributions are summarized as follows:
\begin{itemize}
    \item We propose a novel teacher-student framework to distill textual information from MLLMs and enhance the image fusion process without interfering with inference. A tailored prior distillation loss transfers knowledge from the teacher network to the student, enabling the student network to internalize textual information while reducing model size.
    \item We introduce a spatial-channel cross-fusion module to fully exploit textual information across both spatial and channel dimensions, improving the model's ability to leverage textual priors.
    \item Extensive experiments on multiple datasets demonstrate that our method achieves state-of-the-art (SOTA) performance. Ablation studies further validate the effectiveness of our approach, highlighting its ability to achieve an optimal balance between efficiency and performance while providing valuable insights into model compression and the utilization of prior information.
\end{itemize}

\section{Related Works}
\textbf{Image fusion} has been a well-explored research area, and significant progress has been achieved with the advancements in deep learning. Early methods primarily utilized CNNs~\cite{xu2020u2fusion} to fuse images from different modalities. Subsequently, generative models~\cite{ma2019fusionGAN,zhao2023DDFM} and Transformer-based approaches~\cite{ma2022swinfusion,zhou2024probing} were introduced to enhance fusion quality. Additionally, high-level vision tasks, such as object detection, have been incorporated to guide the fusion process~\cite{cao2023multi}. However, these methods often overlook the degradation present in source images, resulting in suboptimal performance. To address this limitation, recent works, such as Text-IF~\cite{yi2024text} and Text-Fusion~\cite{cheng2023textfusion}, leverage large model priors to achieve degradation-aware fusion. While effective, these approaches introduce substantial computational overhead, limiting their practicality for real-world applications.

\textbf{Knowledge distillation} has gained prominence as a strategy for compressing deep neural networks by transferring knowledge from a large \textquotedblleft teacher\textquotedblright{} model to a smaller \textquotedblleft student\textquotedblright{} model. This approach was initially popularized for reducing the computational overhead of deploying models on resource-constrained devices \cite{hinton2015distilling}. Distillation techniques include output softening, where the student mimics the probability distribution of the teacher, and intermediate-layer guidance, which aligns feature representations \cite{yuan2020revisiting}. In multimodal domains, KD has enabled efficient training of smaller models for tasks involving text and vision, such as CLIP distillation \cite{wu2023tinyclip}. Notably, KD has been explored in text-guided image generation and fusion, where computationally heavy models like Stable Diffusion serve as teachers for lightweight networks \cite{zhao2023mobilediffusion}. However, the distillation of semantic priors from LLMs and multimodal models in the domain of image fusion remains an open problem.

\begin{figure*}[!t]
    \centering
    \includegraphics[width=0.95\linewidth]{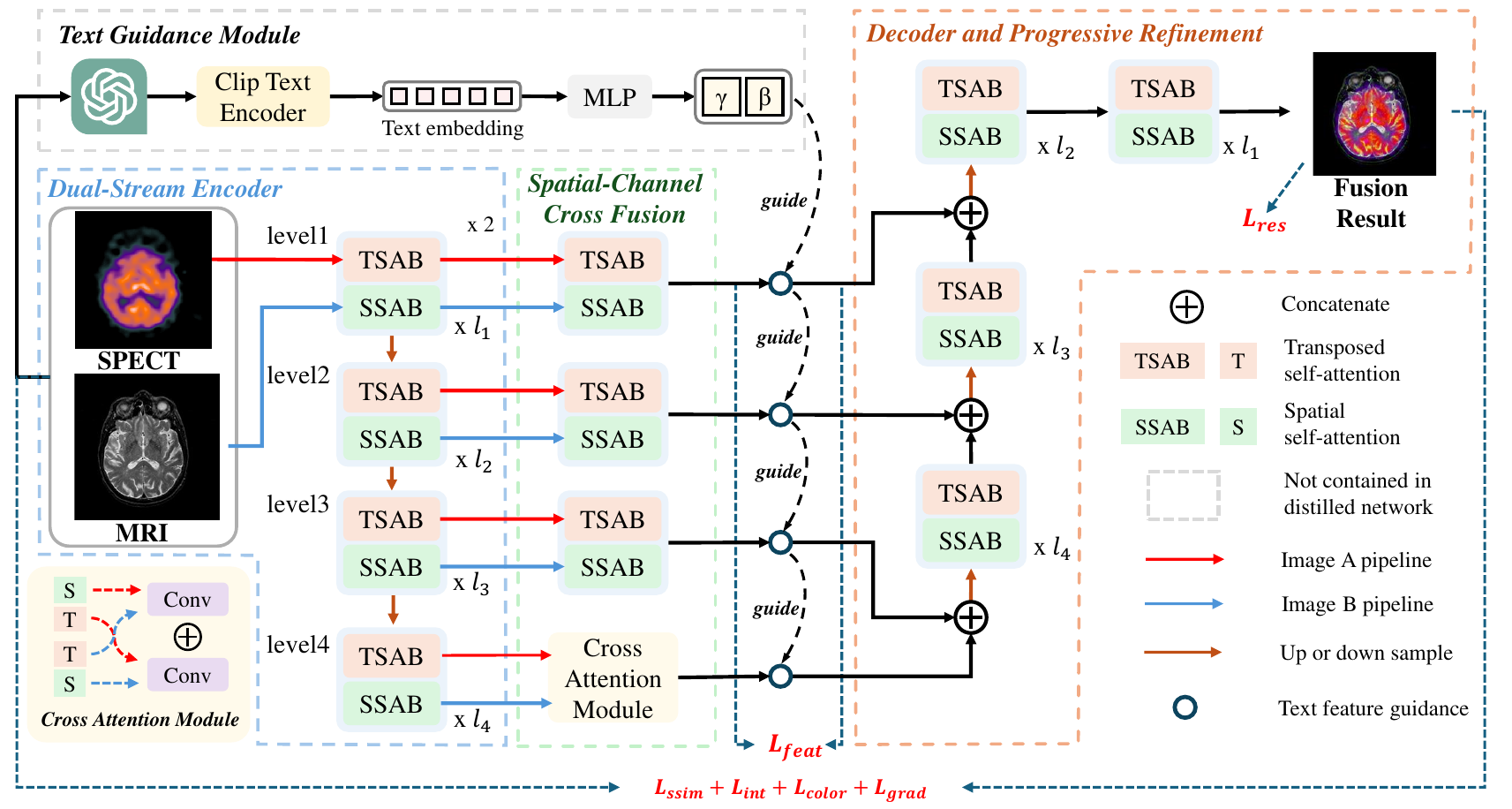}
    \caption{Overview of our text-guided image fusion framework. The architecture consists of three main components: (1) Text Guidance Module that leverages LLMs and CLIP to generate semantic guidance; (2) Encoder that processes visible and infrared inputs through dual-stream transformers with TSAB and SSAB blocks, followed by cross-modal fusion; (3) Decoder and Refinement that progressively reconstructs the fused image with text-guided feature modulation. Gray components are removed in the distilled student network. TSAB: Transposed Self-Attention Block, SSAB: Spatial Self-Attention Block.}
    \label{fig:framework}
\end{figure*}

\section{Method}

\subsection{Framework}
Traditional image fusion methods formulate the task as mapping two source images (e.g., $\mathbf{I_{vis}}$, $\mathbf{I_{ir}}$) to a fused output $\mathbf{I_f}$ through a fusion network $\Phi(.)$:
\begin{equation} \mathbf{I_f} = \Phi(\mathbf{I_{vis}}, \mathbf{I_{ir}}) \end{equation}
Text-guided methods enhance this process by incorporating additional textual input, generated by large language models and encoded by CLIP to produce $\mathbf{T}$. The fused image $\mathbf{I_f}$ is then generated using $\mathbf{T}$ as guidance:
\begin{equation} \mathbf{I_f} = \Phi(\mathbf{I_{vis}}, \mathbf{I_{ir}}, \mathbf{T}) \end{equation}
However, the computational cost of LLMs and CLIP is substantial. To mitigate this, we propose a prior distillation framework consisting of a text-guided teacher network $\Phi_t$ and its distilled student network $\Phi_s$. During training, $\Phi_t$ utilizes textual information from MLLMs, which is then transferred to $\Phi_s$ through our tailored prior distillation loss. Since semantic information is encoded across both spatial and channel dimensions, we introduce spatial-channel cross fusion blocks within the network. As illustrated in Fig.~\ref{fig:framework}, this framework enables the student network to internalize textual information while significantly reducing model size. Both networks share similar structural blocks, with the student network being more compact.

\subsection{Network Architecture}
Our framework consists of two main components: a text-guided teacher network and a lightweight student network. The teacher network serves as a proxy, leveraging priors from the LLM and transferring this information to the student network in the form of features. The student network mimics the feature processing of the teacher, thus acquiring the semantic knowledge indirectly from the LLM. This enables the student network to maintain a small model size while being adaptive to image degradation, without requiring direct text input.

\subsubsection{Text-Guided Teacher Network}
The teacher network adopts a hierarchical structure with four levels of feature processing. Given its critical role in leveraging textual priors, the core operator aims to effectively model features in spatial and channel dimensions~\cite{chen2025comparative}. Additionally, we introduce spatial-channel cross fusion blocks as the core operator, enabling the exploration of correlations between spatial and channel dimensions. Text information is fused through the sequential feature modulation, ensuring effective integration of textual priors.
This network is organized into three primary components:

1) \textbf{Dual-Stream Encoder}: Processes visible and infrared inputs through parallel branches, each consisting of overlapped patch embedding layers and spatial and channel self-attention blocks. The hierarchical feature transformation at level $l$ is expressed as:
\begin{equation} F_{l}^{i} = \mathcal{F}_n \circ \mathcal{F}_{n-1} \circ \cdots \circ \mathcal{F}_1(F_{l-1}^{i}) \end{equation}
where $\mathcal{F}_k$ represents the $k$-th spatial-channel self-attention block for modality $i \in \{\text{vis, ir}\}$, and $\circ$ denotes function composition. Each block integrates sequential Windowed Self-Attention Blocks (SSAB) and Transposed Self-Attention Blocks (TSAB) for comprehensive feature encoding:
\begin{equation} \text{SSAB}(\mathbf{F}) = \mathbf{F} + \text{LN}(\text{MHA}_s(\mathbf{F})) \end{equation} \begin{equation} \text{TSAB}(\mathbf{F}) = \mathbf{F} + \text{LN}(\text{MHA}_c(\mathbf{F})) \end{equation}
Here, $\text{MHA}_c(.)$ and $\text{MHA}_s(.)$ denote multi-head attention mechanisms for channel and spatial dimensions~\cite{Zamir2021Restormer,liu2021swin}, respectively, and LN represents layer normalization.

2) \textbf{Spatial-Channel Cross Fusion}: 
To fully leverage prior information and establish the connection between two modalities, we propose the spatial-channel cross fusion module.
At the upper level, features are concatenated and fused; at the lower level, features from both modalities are fused:
\begin{align} &\mathbf{F}^{\hat{vis}}_{l} = \text{Conv}_{1\times1} \circ C(\text{SSAB}(\mathbf{F}^{vis}_{l}), \text{TSAB}(\mathbf{F}^{ir}_{l})), \\
&\mathbf{F}^{\hat{ir}}_{l} =\text{Conv}_{1\times1} \circ C(\text{SSAB}(\mathbf{F}^{ir}_{l}) ,\text{TSAB}(\mathbf{F}^{vis}_{l})), \\ &\mathbf{F}^{fused}_l = \mathbf{F}^{\hat{vis}}_{l}+\mathbf{F}^{\hat{ir}}_{l}.
\end{align}
where $C(.)$ denotes concatenation. The fused features are then modulated by the text embeddings:
\begin{equation} \hat{\mathbf{F}^{fused}_{l}} = (1 + \gamma(\mathbf{T})) \odot \mathbf{F}^{fused}_{l} + \beta(\mathbf{T}) \end{equation}
where $\gamma(\mathbf{T})$ and $\beta(\mathbf{T})$ are learnable modulation parameters derived from the text embeddings $\mathbf{T}$, and $\odot$ denotes element-wise multiplication.

3) \textbf{Decoder and Progressive Refinement}: The decoder progressively refines features through:
\begin{equation}
    \mathbf{F}_{dec}^l = \mathcal{D}_n \circ \text{Up}(\text{C}(\mathbf{F}_{dec}^{l+1}, \hat{\mathbf{F}_{fused}^l}))
\end{equation}
where $\mathcal{D}_n$ represents a sequence of $n$ decoder blocks similar to encoder. Specially, for the lowest level($l = 1$), we add extra decoder blocks as refinement:
\begin{equation}
    \mathbf{F}_{out} = \mathcal{R}_m \circ \mathcal{R}_{m-1} \circ \ldots \circ \mathcal{R}_1(\mathbf{F}_{dec}^1)
\end{equation}
Each block $\mathcal{R}_k$ consists of SSAB and TSAB operations sharing block number with the encoder at the corresponding level. At the end, the $\mathbf{F}_{out}$ is the final result $\mathbf{I}_{fused}$.

\subsubsection{Lightweight Student Network}
The student network maintains a similar hierarchical structure but achieves 90\% parameter reduction through:
\begin{itemize}
    \item Dimension reduction in transformer blocks (from 48 to 16 channels)
    \item Removal of text guidance module, directly using $\mathbf{F}_{fused}^l$ instead of $\hat{\mathbf{F}_{fused}^l}$
\end{itemize}

\subsection{Prior Knowledge Distillation}
To achieve efficient inference without relying on MLLMs while preserving their semantic prior and reducing model complexity, we propose a two-stage knowledge distillation method to progressively transfer knowledge from MLLMs to a lightweight student network.

\subsubsection{Teacher Network Training}
In the first stage, we transfer knowledge from the MLLM to a teacher network, enabling the teacher network to acquire semantic understanding and degradation-aware fusion capabilities. The teacher network is trained with text guidance using the following loss:
\begin{equation}
\mathcal{L}_{tea} = \lambda_{ssim}^{t} \mathcal{L}_{ssim} + \lambda_{int}^{t} \mathcal{L}_{int} + \lambda_{grad}^{t} \mathcal{L}_{grad} + \lambda_{color}^{t} \mathcal{L}_{color}
\end{equation}
where $\mathcal{L}_{ssim}$ measures structural similarity through SSIM index, $\mathcal{L}_{int}$ preserves maximum intensity information between source images, $\mathcal{L}_{grad}$ ensures gradient consistency using Sobel operators, and $\mathcal{L}_{color}$ maintains color fidelity in YCbCr space. The weights $\lambda^{t}$ are dynamically adjusted based on text guidance. The detailed expressions for each loss component are:
\begin{equation}
\mathcal{L}_{int} = \frac{1}{HW} \|I_f - \max(I_{vis}^g, I_{ir}^g)\|_1
\end{equation}
where $H$ and $W$ are the height and width of the image, $I_f$ is the fused image, and $I_{vis}^g$ and $I_{ir}^g$ represent the visible and infrared guidance images respectively.
\begin{equation}
\mathcal{L}_{SSIM}(t) = (1-SSIM(I_f, I_{vis}^g)) + \delta_{ir}(t)(1-SSIM(I_f, I_{ir}^g))
\end{equation}
where $SSIM(\cdot)$ calculates the structural similarity index, and $\delta_{ir}(t)$ is a text-dependent weight for infrared guidance.
\begin{equation}
\mathcal{L}_{grad} = \frac{1}{HW} \|\nabla I_f - \max(\nabla I_{vis}^g, \nabla I_{ir}^g)\|_1
\end{equation}
where $\nabla$ represents the gradient operator implemented using Sobel filters in both horizontal and vertical directions.
\begin{equation}
\mathcal{L}_{color} = \frac{1}{HW} \|F_{CbCr(I_f)}-F_{CbCr(I_{vis}^{g})}\|_1
\end{equation}
where $F_{CbCr}$ denotes the transfer function of RGB to YCbCr.

These loss terms work together to ensure high-quality image fusion: $\mathcal{L}_{int}$ maintains the maximum intensity information from both source images, $\mathcal{L}_{SSIM}$ preserves structural details through the SSIM~\cite{SSIM} metric, $\mathcal{L}_{grad}$ ensures edge consistency through gradient preservation, and $\mathcal{L}_{color}$ maintains natural color appearance. The text-dependent weights $\lambda^{t}$ allow dynamic adjustment of each loss component's contribution based on the specific fusion requirements described in the text prompt.

\subsubsection{Progressive Knowledge Transfer}
In the second stage, the student network is trained to progressively acquire knowledge from the teacher network by mimicking its feature processing through three complementary loss functions:

1) Feature consistency loss:
\begin{equation}
\mathcal{L}_{feat} = \sum_{l=1}^L \| \text{Down}(F_{t}^l) - F_{s}^l\|_1
\end{equation}
where Down represents learnable dimension reduction to match teacher-student feature dimensions.

2) Output reconstruction loss:
\begin{equation}
    \mathcal{L}_{res} = \| I^{fused}_t - I^{fused}_s \|_1
\end{equation}

3) Base fusion loss $\mathcal{L}_{base}$ inherited from teacher but with fixed weights.

The final distillation objective is:
\begin{equation}
\mathcal{L}_{distill} = \alpha_1\mathcal{L}_{base} + \alpha_2\mathcal{L}_{feat} + \alpha_3 \mathcal{L}_{res}
\end{equation}
This progressive distillation strategy enables knowledge transfer while reducing computational efficiency.

\section{Experiments}

We conduct extensive experiments to evaluate our proposed method on both infrared-visible fusion (IVF) and medical image fusion tasks. In this section, we first introduce the implementation details and datasets, then present comprehensive comparisons with state-of-the-art methods, followed by detailed ablation studies.

\subsection{Implementation Details and Datasets}

\textbf{Implementation Details}: Both teacher and student networks are trained using AdamW optimizer with a learning rate of 0.0001. Input images are cropped to $128 \times 128$ patches during training. For the teacher network, the hyper-parameters $\{\lambda_{int}, \lambda_{ssim}, \lambda_{grad}, \lambda_{color}\}$ are set to $\{24, 40, 48, 12\}$ respectively. When text guidance is enabled, task-specific parameters are provided in the supplementary material. We employ QWen2VL \cite{Qwen2VL}, an open-source large vision-language model, to generate descriptive text labels for input image pairs. 
All experiments are conducted on four NVIDIA GeForce RTX 4090 GPUs using PyTorch framework.

\textbf{Datasets}: For IVF tasks, we evaluate on three widely-used datasets: MSRS~\cite{MSRS_dataset}, M3FD \cite{liu2022target}, and RoadScene \cite{roadscene}. These datasets cover diverse scenarios including urban scenes, indoor environments, and road conditions with varying lighting and weather conditions. We use the training split of MSRS for training and evaluate on its test split as well as the entire M3FD and RoadScene datasets. For medical image fusion, we utilize the Harvard Medical Image Fusion Dataset, which contains three modality pairs: SPECT-MRI, CT-MRI, and PET-MRI. Each modality pair is trained and evaluated separately.

\textbf{Evaluation Metrics}: For IVF tasks, we employ three complementary metrics: Information Entropy (EN) \cite{EN} to quantify information density, Visual Information Fidelity (VIF) \cite{VIF} to evaluate perceptual quality, and Quality of Gradient-based Fusion ($Q^{AB/F}$) \cite{Qabf} to assess edge preservation. For medical image fusion, we include Structural Similarity (SSIM) \cite{SSIM} due to its critical role in medical applications, which is calculated as the sum of the SSIM values between each source image and the fused image. The metrics are chosen for their strong correlation with the source images, as they take all of the images as input. 

\textbf{Benchmark Methods}: We compare our teacher and distilled networks with several state-of-the-art methods, including SwinFusion~\cite{ma2022swinfusion}, U2Fusion~\cite{xu2020u2fusion}, CDDFuse~\cite{zhao2023cddfuse}, FusionGAN~\cite{ma2019fusionGAN}, PIAFusion~\cite{tang2022piafu}, SuperFusion~\cite{TANG2022SuperFusion}, IFCNN~\cite{zhang2020IFCNN}, DDFM~\cite{zhao2023DDFM}, Text-IF~\cite{yi2024text}, TC-MOA~\cite{zhu2024tcmoa}, PSLPT~\cite{wang2024pslpt}, EMFusion~\cite{xu2021emfusion}, Zero~\cite{zero}, and MSRPAN~\cite{MSRPAN}.
These methods include transformer-based, unsupervised learning, dual-branch, generative models, illumination- and semantic-aware fusion, CNN-based fusion, diffusion models, and text-guided fusion (Text-IF). Beyond performance comparisons, we focus on analyzing the impact of different components and parameters on both model effectiveness and inference efficiency through comprehensive ablation studies.

\subsection{Comparison on IVF Tasks}

Table \ref{tab:ivf_comparison_big} presents the quantitative comparison with SOTA methods on three IVF datasets. Our teacher network consistently outperforms existing methods across all metrics and datasets, demonstrating the effectiveness of text-guided fusion. On the MSRS dataset, our teacher network achieves significant improvements in $Q^{AB/F}$ compared to the previous best method Text-IF.
\begin{figure}[ht] 
    \centering
    \includegraphics[width=\linewidth]{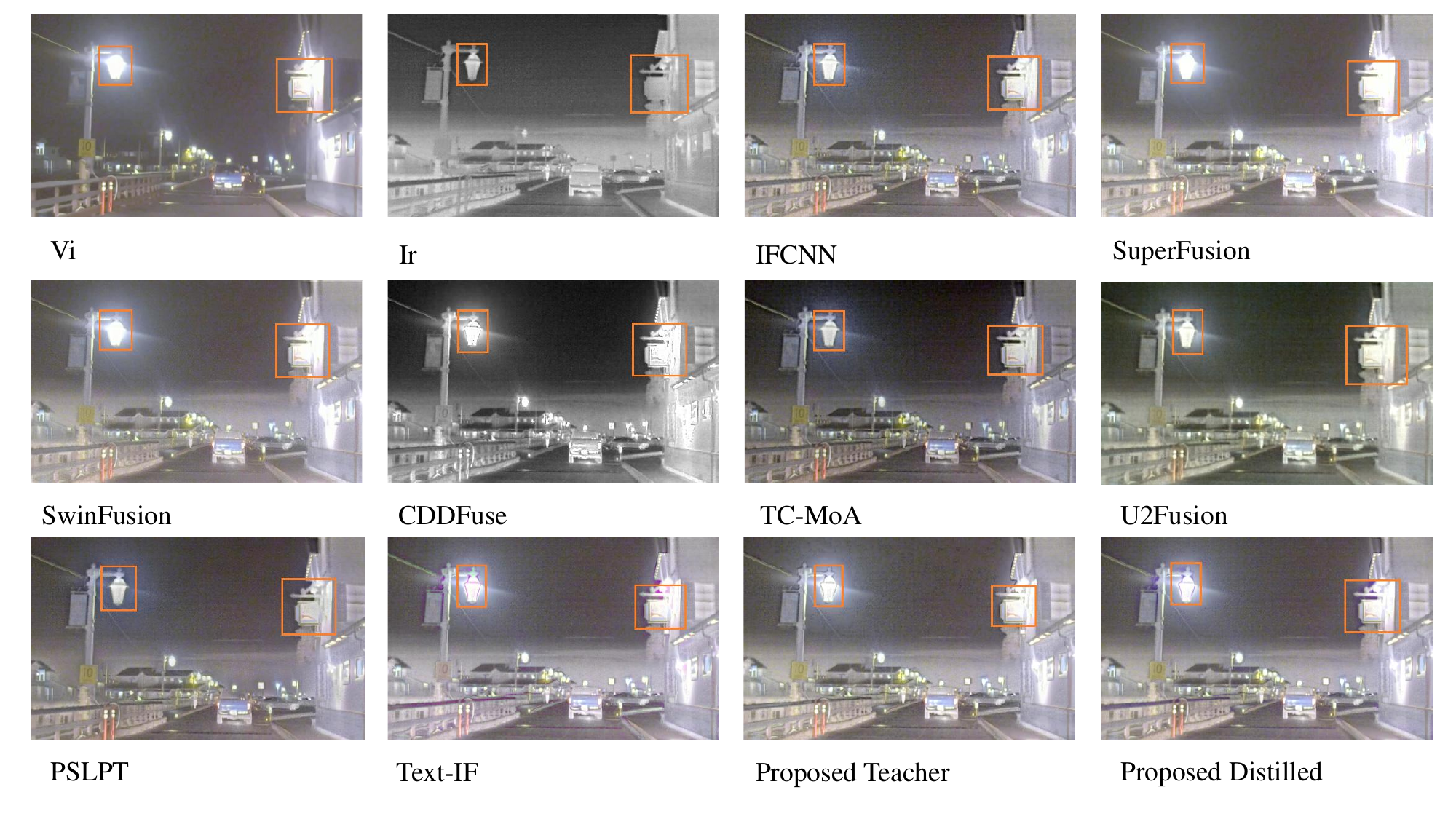} 
    \caption{Qualitative comparison of different image fusion methods on a challenging scene (FLIR\_05767.jpg) from the RoadScene dataset. Our method better preserves thermal information from infrared images while maintaining visible details and natural appearance, especially in scenarios with extreme lighting conditions or complex textures. For a more exhaustive visual comparison across various scenarios and methods, please refer to Fig.~\ref{fig:ivf_compare_full}.}
    \label{fig:ivf_compare}
\end{figure}
More remarkably, our distilled student network not only maintains but sometimes exceeds the teacher's performance, particularly in EN and $Q^{AB/F}$ metrics, while reducing parameters by 90\%. This suggests that the knowledge distillation process effectively transfers the teacher's knowledge to a much smaller model. The slight performance variations between teacher and student networks can be attributed to the extended training during the distillation process, where the student benefits from both the teacher's guidance and task-specific objectives. Another reason is that current domain models are generally over-parameterized compared to relatively small datasets, which is detailed in the ablation experiments.

\begin{table*}[!t]
    \centering
    \caption{Quantitative comparison with SOTA methods on IVF tasks across MSRS, M3FD, and RoadScene datasets. \textbf{Bold} and \underline{underlined} values indicate the best and second-best results respectively. Tra.P. = Trainable Parameters, Fix.P. = Fixed Parameters.}
    \label{tab:ivf_comparison_big}
    \begin{tabular}{l|c|cc||ccccc}
        \toprule
        Dataset & Method & Tra.P. (M) & Fix.P. (M) & EN$\uparrow$ & MI$\uparrow$ & SF$\uparrow$ & VIF$\uparrow$ & $Q^{AB/F}\uparrow$ \\
        \midrule
        \multirow{10}{*}{MSRS} 
        & SuperFusion~\cite{TANG2022SuperFusion} & 11.23 & - & 6.587 & 3.596 & 10.783 & 0.813 & 0.557 \\
        & CDDFuse~\cite{zhao2023cddfuse} & 2.40 & - & 6.701 & 3.657 & 12.083 & 0.819 & 0.548 \\
        & IFCNN~\cite{zhang2020IFCNN} & 0.08 & - & 5.975 & 1.706 & 12.734 & 0.579 & 0.479 \\
        & U2Fusion~\cite{xu2020u2fusion} & 0.63 & - & 5.246 & 2.183 & 9.242 & 0.512 & 0.391 \\
        & SwinFusion~\cite{ma2022swinfusion} & 13.04 & - & 6.619 & 3.652 & 11.038 & 0.825 & 0.558 \\
        & PSLPT~\cite{wang2024pslpt} & 3.06 & - & 6.307 & 2.284 & 10.419 & 0.753 & 0.553 \\
        & TC-MOA~\cite{zhu2024tcmoa} & 7.24 & (+329) & 6.633 & 3.251 & 9.370 & 0.811 & 0.565 \\
        & Text-IF~\cite{yi2024text} & 63.8 & (+151) & 6.729 & \textbf{5.406} & 17.384 & 1.051 & 0.690 \\
        & Ours-teacher & 40.3 & (+151) & \underline{6.749} & \underline{4.883} & \underline{17.914} & \underline{1.060} & \underline{0.732} \\
        & Ours-distilled & 4.10 & - & \textbf{6.763} & 4.867 & \textbf{18.299} & \textbf{1.075} & \textbf{0.734} \\
        \midrule
        \multirow{10}{*}{M3FD}
        & SuperFusion~\cite{TANG2022SuperFusion} & 11.23 & - & 6.726 & 4.345 & 11.748 & 0.664 & 0.522 \\
        & CDDFuse~\cite{zhao2023cddfuse} & 2.40 & - & \underline{7.070} & 3.994 & 17.578 & 0.802 & 0.613 \\
        & IFCNN~\cite{zhang2020IFCNN} & 0.08 & - & 6.935 & 2.630 & 16.250 & 0.685 & 0.590 \\
        & U2Fusion~\cite{xu2020u2fusion} & 0.63 & - & 6.872 & 2.683 & 14.248 & 0.673 & 0.578 \\
        & SwinFusion~\cite{ma2022swinfusion} & 13.04 & - & 6.844 & 4.020 & 14.415 & 0.746 & 0.616 \\
        & PSLPT~\cite{wang2024pslpt} & 3.06 & - & \textbf{7.204} & 4.563 & 6.439 & \textbf{0.958} & 0.321 \\
        & TC-MOA~\cite{zhu2024tcmoa} & 7.24 & (+329) & 6.747 & 2.856 & 11.221 & 0.579 & 0.508 \\
        & Text-IF~\cite{yi2024text} & 63.8 & (+151) & 6.849 & \textbf{5.553} & 14.484 & 0.780 & 0.550 \\
        & Ours-teacher & 40.3 & (+151) & 6.965 & 4.780 & \underline{17.949} & 0.896 & \textbf{0.706} \\
        & Ours-distilled & 4.10 & - & 6.989 & \underline{4.898} & \textbf{18.507} & \underline{0.927} & \underline{0.704} \\
        \midrule
        \multirow{10}{*}{RoadScene}
        & SuperFusion~\cite{TANG2022SuperFusion} & 11.23 & - & 6.990 & \underline{3.562} & 12.185 & 0.608 & 0.452 \\
        & CDDFuse~\cite{zhao2023cddfuse} & 2.40 & - & \textbf{7.475} & 3.001 & 19.779 & 0.610 & 0.450 \\
        & IFCNN~\cite{zhang2020IFCNN} & 0.08 & - & 7.222 & 2.842 & 15.998 & 0.591 & 0.536 \\
        & U2Fusion~\cite{xu2020u2fusion} & 0.63 & - & 6.739 & 2.578 & 15.282 & 0.564 & 0.506 \\
        & SwinFusion~\cite{ma2022swinfusion} & 13.04 & - & 7.000 & 3.334 & 12.161 & 0.614 & 0.450 \\
        & PSLPT~\cite{wang2024pslpt} & 3.06 & - & 7.077 & 2.001 & 9.172 & 0.134 & 0.171 \\
        & TC-MOA~\cite{zhu2024tcmoa} & 7.24 & (+329) & \underline{7.387} & 2.853 & 12.786 & 0.577 & 0.477 \\
        & Text-IF~\cite{yi2024text} & 63.8 & (+151) & 7.332 & \textbf{5.009} & 14.199 & 0.739 & 0.578 \\
        & Ours-teacher & 40.3 & (+151) & 7.248 & 3.454 & \textbf{20.891} & \underline{0.743} & \textbf{0.639} \\
        & Ours-distilled & 4.10 & - & 7.279 & 3.328 & \underline{20.082} & \textbf{0.751} & \underline{0.634} \\
        \bottomrule
    \end{tabular}
\end{table*}

\subsection{Comparison on Medical Datasets}
To demonstrate the generalization capability of our method, we evaluate its performance on medical image fusion tasks. As shown in Table \ref{tab:medical_comparison_big}, both our teacher and student networks achieve superior performance across different modality pairs. The improvement is particularly significant in structural preservation metrics (SSIM and $Q^{AB/F}$), which is crucial for medical applications.
\begin{figure*}[!t] 
    \centering
    \includegraphics[width=\linewidth]{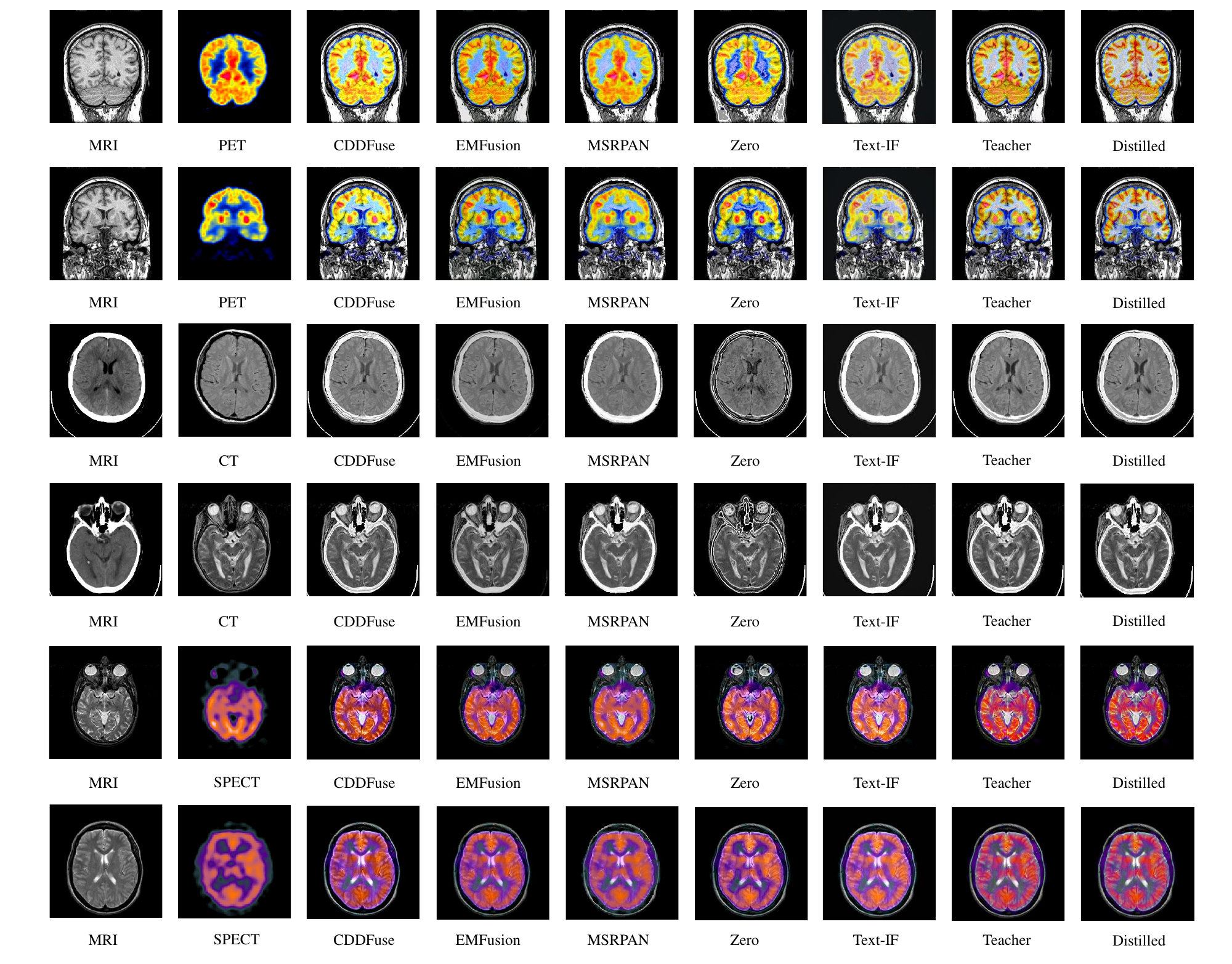} 
    \caption{Comprehensive visual comparison of different image fusion methods on Harvard Medical Image Fusion Datasets (PET-MRI, CT-MRI, SPECT-MRI). For each set of results, from left to right: Modality 1 Input, Modality 2 Input, Text-IF Output, Ours-Teacher Output, Ours-Distilled Output.}
    \label{fig:med_compare_full} 
\end{figure*}
For SPECT-MRI fusion, our method shows notable advantages in preserving functional information while maintaining anatomical details. In CT-MRI fusion, where structural alignment is critical, our approach achieves the highest SSIM scores, indicating better structural preservation. The PET-MRI results further confirm our method's effectiveness in handling multi-modal medical images with different characteristics.
Fig.~\ref{fig:med_compare_full} presents visual examples from different medical modalities. Our fusion results exhibit better detail preservation and contrast enhancement, which is essential for clinical applications. The distilled student network maintains these advantages while significantly reducing computational requirements, making it more practical for clinical deployment.

\begin{table*}[!b]
    \centering
    \caption{Quantitative comparison with SOTA methods on medical image fusion tasks. \textbf{Bold} and \underline{underlined} values indicate the best and second-best results respectively. The additional parameters (in parentheses) are non-trainable but used during inference.}
    \label{tab:medical_comparison_big}
    \begin{tabular}{l|c|c||ccccc}
        \toprule
        Dataset & Method & Parameters (M) & SSIM$\uparrow$ & VIF$\uparrow$ & $Q^{AB/F}\uparrow$ & MI$\uparrow$ & EN$\uparrow$ \\
        \midrule
        \multirow{10}{*}{PET-MRI} 
        & PSLPT~\cite{wang2024pslpt} & 3.06 & 0.815 & 0.548 & 0.373 & 2.641 & 5.492 \\
        & EMFusion~\cite{xu2021emfusion} & 0.78 & 1.221 & 0.685 & 0.783 & 3.207 & \textbf{5.646} \\
        & MSRPAN~\cite{MSRPAN} & 0.39 & 1.182 & 0.581 & \textbf{0.799} & 4.119 & 5.073 \\
        & SwinFusion~\cite{ma2022swinfusion} & 13.04 & 0.725 & 0.703 & 0.683 & 3.662 & 5.817 \\
        & Zero~\cite{zero} & 19.1 & 1.162 & 0.635 & 0.774 & 3.786 & 5.495 \\
        & U2Fusion~\cite{xu2020u2fusion} & 0.63 & 0.494 & 0.460 & 0.292 & 2.785 & 5.532 \\
        & CDDFuse~\cite{zhao2023cddfuse} & 2.40 & 1.227 & 0.650 & 0.765 & 3.572 & 5.149 \\
        & Text-IF~\cite{yi2024text} & 63.8 (+151) & \underline{1.232} & 0.640 & 0.690 & 3.718 & 5.443 \\
        & Ours-teacher & 40.3 (+151) & 1.223 & \underline{0.909} & 0.782 & \underline{4.248} & \underline{5.611} \\
        & Ours-distilled & 4.10 & \textbf{1.243} & \textbf{0.929} & \underline{0.784} & \textbf{4.318} & 5.474 \\
        \midrule
        \multirow{10}{*}{CT-MRI}
        & PSLPT~\cite{wang2024pslpt} & 3.06 & 0.810 & 0.502 & 0.432 & 2.392 & 4.730 \\
        & EMFusion~\cite{xu2021emfusion} & 0.78 & 1.266 & 0.552 & 0.475 & 3.116 & 4.785 \\
        & MSRPAN~\cite{MSRPAN} & 0.39 & 1.261 & 0.436 & 0.455 & \textbf{4.126} & 4.202 \\
        & SwinFusion~\cite{ma2022swinfusion} & 13.04 & 0.579 & 0.522 & 0.545 & 3.190 & \underline{5.144} \\
        & Zero~\cite{zero} & 19.1 & 1.199 & 0.320 & \underline{0.582} & 3.358 & 4.406 \\
        & U2Fusion~\cite{xu2020u2fusion} & 0.63 & 0.042 & 0.074 & 0.489 & 1.694 & 4.896 \\
        & CDDFuse~\cite{zhao2023cddfuse} & 2.40 & 1.224 & 0.526 & 0.530 & \underline{3.683} & \textbf{5.733} \\
        & Text-IF~\cite{yi2024text} & 63.8 (+151) & \textbf{1.313} & 0.542 & 0.561 & 3.211 & 4.356 \\
        & Ours-teacher & 40.3 (+151) & \textbf{1.313} & \underline{0.641} & \textbf{0.657} & 3.246 & 4.462 \\
        & Ours-distilled & 4.10 & \underline{1.312} & \textbf{0.653} & \textbf{0.657} & 3.233 & 4.494 \\
        \midrule
        \multirow{10}{*}{SPECT-MRI}
        & PSLPT~\cite{wang2024pslpt} & 3.06 & 0.933 & 0.359 & 0.325 & 2.747 & 5.140 \\
        & EMFusion~\cite{xu2021emfusion} & 0.78 & \textbf{1.212} & 0.665 & 0.692 & 3.210 & 4.911 \\
        & MSRPAN~\cite{MSRPAN} & 0.39 & 1.153 & 0.525 & 0.560 & \underline{4.334} & 4.753 \\
        & SwinFusion~\cite{ma2022swinfusion} & 13.04 & 0.684 & 0.744 & 0.720 & 3.795 & \textbf{5.401} \\
        & Zero~\cite{zero} & 19.1 & 1.180 & 0.582 & 0.681 & 3.564 & 4.997 \\
        & U2Fusion~\cite{xu2020u2fusion} & 0.63 & 0.479 & 0.419 & 0.696 & 2.870 & 4.539 \\
        & CDDFuse~\cite{zhao2023cddfuse} & 2.40 & 1.169 & 0.786 & 0.719 & 4.109 & 4.396 \\
        & Text-IF~\cite{yi2024text} & 63.8 (+151) & 1.200 & 0.747 & 0.715 & 3.994 & 4.807 \\
        & Ours-teacher & 40.3 (+151) & \underline{1.210} & \textbf{0.888} & \underline{0.746} & 4.248 & 5.035 \\
        & Ours-distilled & 4.10 & 1.202 & \underline{0.887} & \textbf{0.747} & \textbf{4.371} & \underline{5.361} \\
        \bottomrule
    \end{tabular}
\end{table*}

\subsection{Ablation Study}

To thoroughly evaluate the effectiveness of our proposed method, we conduct comprehensive ablation studies on both the teacher and student networks using the MSRS test set. These experiments examine the impact of text guidance, various loss components, and model architectures. Specifically, we address key questions such as: Why does the distilled network sometimes achieve a higher score than the teacher? Is it more effective to train a small model directly instead of using distillation?

\begin{figure}[!t]
    \centering
    \includegraphics[width=\linewidth]{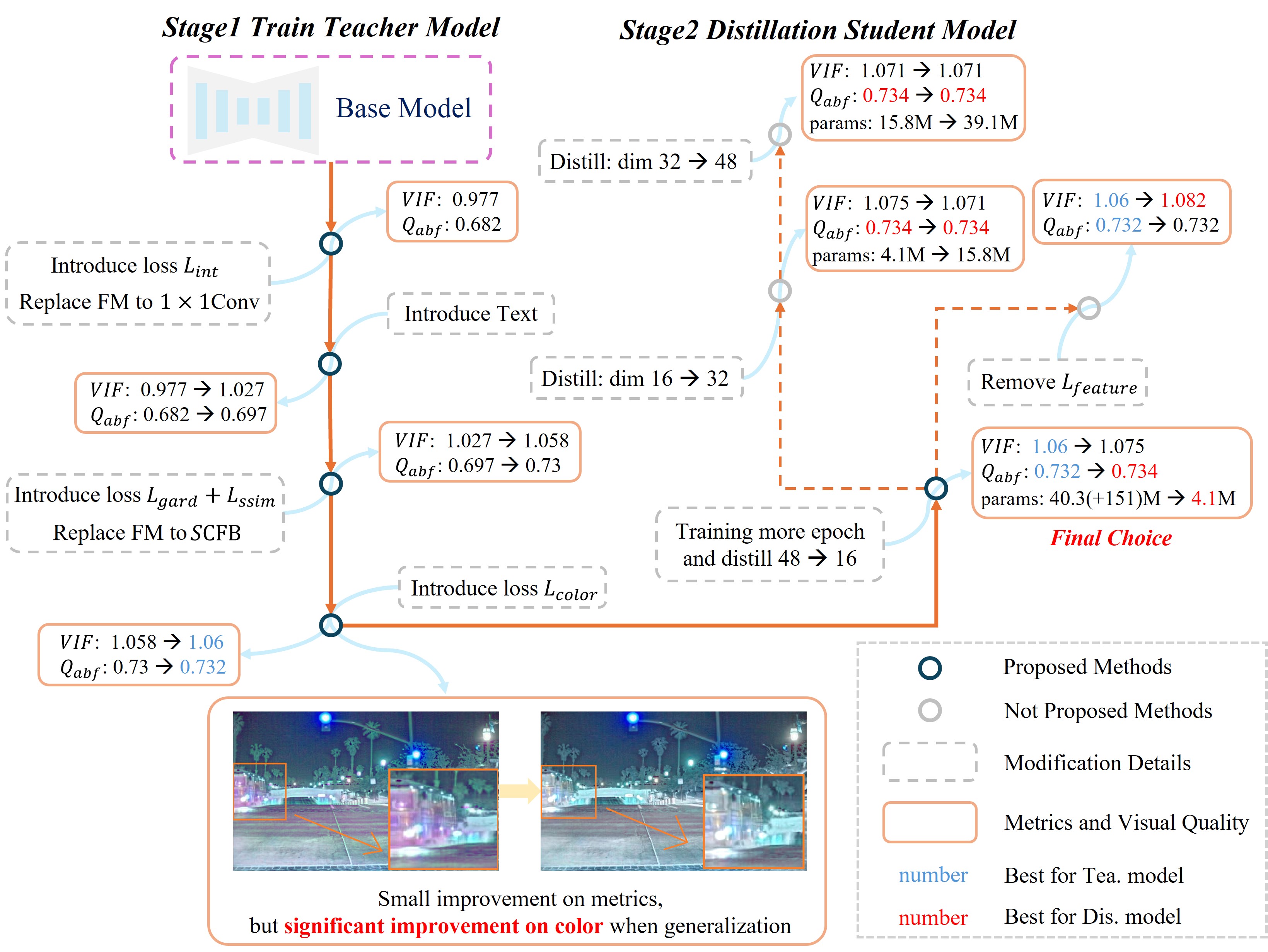}
    \caption{Overview of our ablation study.}
    \label{fig:ablation}
\end{figure}

\subsubsection{Analysis of Teacher Network}

\paragraph{Effect of Text Guidance.} To validate the effectiveness of text guidance, we compare our full model with its variant without text information. Removing text guidance leads to a significant performance drop across all metrics (EN: -0.023, VIF: -0.083, $Q^{AB/F}$: -0.05). This demonstrates that text descriptions provide valuable semantic information that helps the model better understand and fuse image features. The text guidance acts as an advanced form of reference information, offering rich contextual information beyond pixel-level features.

\paragraph{Impact of Different Loss Components.}  
We evaluate the contribution of each loss component by systematically removing them from the full model. The results, as shown in Table~\ref{tab:ablation_loss_teacher}, indicate that all loss components positively impact the final performance, although their importance varies. The intensity loss ($\mathcal{L}_{int}$) and gradient loss ($\mathcal{L}_{grad}$) have the most significant influence, highlighting their essential role in preserving structural information. The SSIM loss ($\mathcal{L}_{ssim}$) aids in maintaining perceptual quality, while the color loss ($\mathcal{L}_{color}$) ensures the natural appearance of fused results. 

Due to the nature of our method, which reconstructs the entire image rather than only modifying the Y-axis in YCbCr color space, $\mathcal{L}_{color}$ plays a particularly important role. Although most metrics convert images to grayscale for evaluation, making $\mathcal{L}_{color}$ appear to have minimal effect on metric scores, it is critical for producing visually appealing results in the final fused images. 
\begin{table}[h]
    \centering
    \caption{Ablation study on loss components for the teacher network. Results are reported on the MSRS test set.}
    \label{tab:ablation_loss_teacher}
    \begin{tabular}{l|ccc}
        \toprule
        Method & EN$\uparrow$ & VIF$\uparrow$ & $Q^{AB/F}\uparrow$ \\
        \midrule
        $\mathcal{L}_{int}$ & 6.695 & 1.027 & 0.697 \\
        $\mathcal{L}_{int}+\mathcal{L}_{color}$ & 6.696 & 1.025 & 0.698 \\
        $\mathcal{L}_{int}+\mathcal{L}_{grad}+\mathcal{L}_{ssim}$ & 6.731 & 1.058 & 0.730 \\
        $\mathcal{L}_{int}+\mathcal{L}_{color}+\mathcal{L}_{grad}$ & 6.733 & 1.059 & 0.733 \\
        $\mathcal{L}_{color}+\mathcal{L}_{grad}+\mathcal{L}_{ssim}$ & 6.676 & 1.063 & 0.728 \\
        $\mathcal{L}_{color}+\mathcal{L}_{grad}+\mathcal{L}_{ssim}+\mathcal{L}_{int}$ & \textbf{6.763} & \textbf{1.075} & \textbf{0.734} \\
        \bottomrule
    \end{tabular}
\end{table}

\paragraph{Feature Fusion Methods.} We investigate different fusion embedding strategies to combine features from infrared and visible images. The comparison includes concatenation \& 1$\times$1 conv, unlearnable weighted attention, dynamic weighted conv, multiscale conv, and our proposed adaptive fusion method. As shown in Table~\ref{tab:ablation_fusion}, our adaptive fusion approach achieves superior performance by dynamically adjusting the fusion weights based on spatial and channel features.
\begin{table}[h]
    \centering
    \caption{Comparison of different fusion modules. Results on MSRS test set.}
    \label{tab:ablation_fusion}
    \begin{tabular}{l|ccc}
        \toprule
        Method & EN$\uparrow$ & VIF$\uparrow$ & $Q^{AB/F}\uparrow$ \\
        \midrule
        Cat \& 1$\times$1 conv & 6.738 & 1.059 & 0.728 \\
        Unlearn weight atten & 6.667 & 0.971 & 0.655 \\
        Dynamic weight conv & 6.737 & 1.058 & 0.729 \\
        Multiscale conv & 6.743 & 1.053 & 0.720 \\
        Ours & \textbf{6.749} & \textbf{1.060} & \textbf{0.732} \\
        \bottomrule
    \end{tabular}
\end{table}

\subsubsection{Analysis of Student Network}

\paragraph{Effect of Model Size.} A notable finding in our distillation experiments is that reducing the model size by up to 90\% (from 40.3M to 4.1M parameters if not calculate CLIP) does not significantly impact performance. As shown in Table~\ref{tab:ablation_size}, the compact student model maintains comparable or even slightly better results across different metrics. This suggests that the original model might be over-parameterized for the current fusion task, and the knowledge distillation process effectively transfers essential fusion capabilities to a much smaller architecture.
\begin{table}[h]
    \centering
    \caption{Impact of model size on distilled network. Results on MSRS test set.}
    \label{tab:ablation_size}
    \begin{tabular}{l|c|ccc}
        \toprule
        Model & Params(M) & EN$\uparrow$ & VIF$\uparrow$ & $Q^{AB/F}\uparrow$ \\
        \midrule
        48-dim & 39.1 & \textbf{6.780} & 1.071 & \textbf{0.734} \\
        32-dim & 15.8 & 6.768 & 1.071 & \textbf{0.734} \\
        16-dim & \textbf{4.1} & 6.763 & \textbf{1.075} & \textbf{0.734} \\
        \bottomrule
    \end{tabular}
\end{table}

\paragraph{Impact of Distillation Losses.} 
We evaluate the contribution of different components in our distillation framework. The results in Table~\ref{tab:ablation_loss_distill} show that result-level supervision (matching the teacher’s output) provides strong guidance, significantly improving the student model’s performance. The feature distillation loss, on the other hand, plays a key role in preserving feature-level consistency, ensuring that the student model captures important intermediate representations from the teacher. By introducing feature distillation, the model learns high-frequency details and intermediate features, enriching the information. This improves metrics like $EN$ and $Q^{AB/F}$, though it may slightly reduce the $VIF$ score due to imperfect alignment with perceptual fidelity.
\begin{table}[h]
    \centering
    \caption{Ablation study on distillation losses (MSRS dataset).}
    \label{tab:ablation_loss_distill}
    \begin{tabular}{l|ccc}
        \toprule
        Method & EN$\uparrow$ & VIF$\uparrow$ & $Q^{AB/F}\uparrow$ \\
        \midrule
        $\mathcal{L}_{base}$ & 6.728 & 0.989 & 0.681 \\
        $\mathcal{L}_{base}+\mathcal{L}_{res}$ & 6.761 & \textbf{1.082} & \underline{0.732} \\
        $\mathcal{L}_{base}+\mathcal{L}_{feat}+\mathcal{L}_{res}$ & \textbf{6.763} & \underline{1.075} & \textbf{0.734} \\
        \bottomrule
    \end{tabular}
\end{table}

\begin{figure}[!t]
    \centering
    \includegraphics[width=\linewidth]{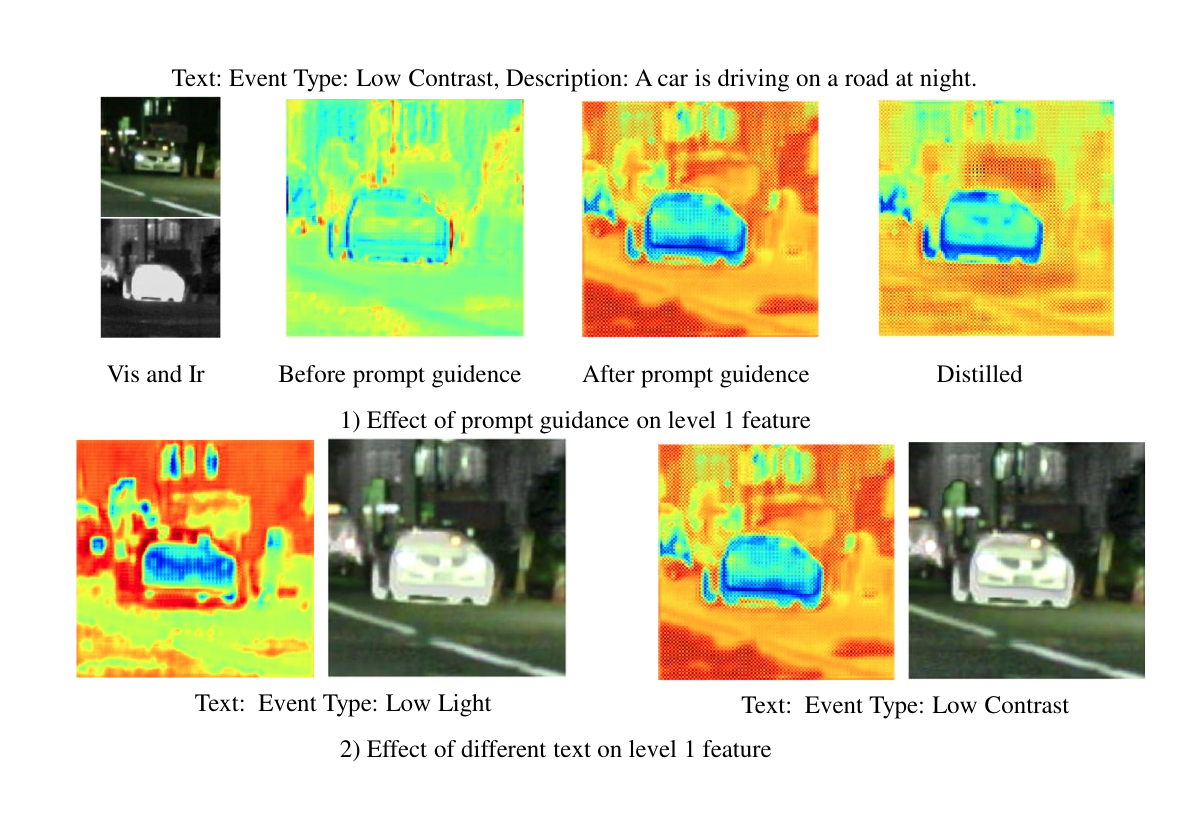}
    \caption{Effect of text guidance on level 1 and fused image. (a) Original visible and infrared input images. (b) Feature maps before text guidance. (c) Feature maps after text guidance. (d) Final fused images under different text descriptions (``Low Contrast'' and ``Low Light''). (e) Distilled student network's output.}
    \label{fig:text}
\end{figure}

\paragraph{Text Prior Analysis.}
To better understand how textual priors influence the fusion process and how they are distilled, we conducted a detailed analysis of their effects at different network levels. As shown in Fig.~\ref{fig:text}, text guidance significantly impacts the model's attention to different image regions. For instance, when processing a low-contrast nighttime scene, the text prior helps the model better distinguish the vehicle's contours from the background, evident in the feature maps showing more defined object boundaries after prompt guidance. Different text descriptions (e.g., ``Low Contrast'' vs. ``Low Light'') lead to distinct feature emphasis patterns. The distilled results show that our student network successfully inherits these text-guided attention patterns without explicit textual input, validating the effectiveness of our prior distillation approach in preserving semantic understanding.

\begin{figure}[!t]
    \centering
    \includegraphics[width=\linewidth]{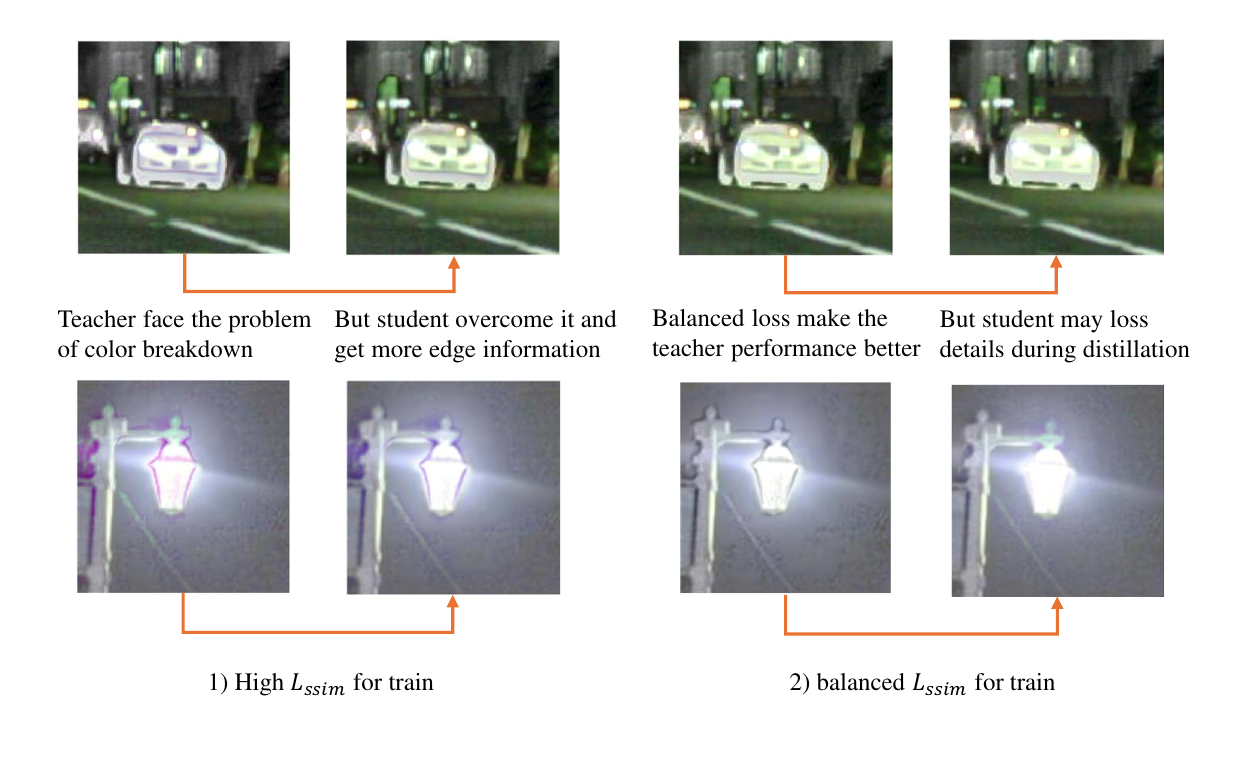}
    \caption{When teacher performs extremely, the student learns more information. This figure illustrates how a student network learns to balance structural and color information even when the teacher over-prioritizes structural details, leading to color breakdown in the teacher's output.}
    \label{fig:kd}
\end{figure}

\paragraph{Knowledge Distillation Analysis.}
An intriguing observation from our experiments is that the distilled student network occasionally outperforms its teacher model. This phenomenon, illustrated in Fig.~\ref{fig:kd}, suggests that the distillation process can act as a form of regularization. When training the teacher network with an extremely high weight for $\mathcal{L}_{ssim}$ (10$\times$ the baseline), the teacher becomes highly specialized in edge detection, sacrificing color fidelity. Surprisingly, the student, when distilled from such a teacher, learns to balance these competing objectives, maintaining edge clarity while preserving natural color representation. This indicates that extreme specialization in the teacher can provide a clearer supervision signal for certain features, enabling the student to learn more comprehensive fusion strategies. In our final implementation, a moderately increased weight for $\mathcal{L}_{ssim}$ proves sufficient for the student to achieve superior performance across multiple metrics with stable training.

Our ablation studies reveal several key insights:
\begin{itemize}
    \item Text guidance significantly enhances fusion performance by embedding rich semantic context into the process.
    \item The integration of multiple loss terms is crucial for achieving balanced and robust fusion results.
    \item Model compression via knowledge distillation achieves comparable performance with drastically fewer parameters, as the teacher model and some text-guided methods exhibit excessive over-parameterization compared to small datasets.
\end{itemize}

\subsection{Inference Time Analysis}

To evaluate the computational efficiency, we analyze the inference times of different components across the previous state-of-the-art (SoTA) text-guided method Text-IF, our teacher network, and the distilled student network. For text generation, we use Qwen2-VL~\cite{Qwen2VL} as the large language model (LLM), which operates on two NVIDIA RTX 4090 GPUs with the fast attention library~\cite{dao2023flashattention2}. Other timing experiments were conducted on a separate single NVIDIA RTX 4090 GPU, running with a batch size of 1 and an image resolution of $128 \times 128$ pixels. Since these results may vary across different machines or even different states of the same machine, this comparison should be considered non-rigorous.

As shown in Table~\ref{tab:inference_time}, the LLM component dominates the computational cost, requiring over 2 seconds per image. CLIP encoding adds another 110-115ms overhead. While the teacher network's fusion module takes 133.4ms, our distilled student network reduces this to just 46.1ms---a 65\% reduction in fusion time. More importantly, by eliminating the need for LLM and CLIP during inference, our student network achieves a dramatic 98\% reduction in total inference time (from 2556.91ms to 46.11ms).

\begin{table}[ht]
    \centering
    \caption{Detailed inference time comparison (in milliseconds) for different components. LLM refers to the Qwen2-VL model used for text generation.}
    \label{tab:inference_time}
    \begin{tabular}{l|cccc|c}
        \toprule
        Method & Data Load & LLM & CLIP & Fusion & Total \\
        \midrule
        Text-IF & 0.01 & 2261.2 & 115.7 & 152.1 & 2529.01 \\
        Teacher & 0.01 & 2311.1 & 112.4 & 133.4 & 2556.91 \\
        Student & 0.01 & - & - & \textbf{46.1} & 46.11 \\
        \bottomrule
    \end{tabular}
\end{table}

\begin{figure*}[t!]
    \centering
    \includegraphics[width=\linewidth]{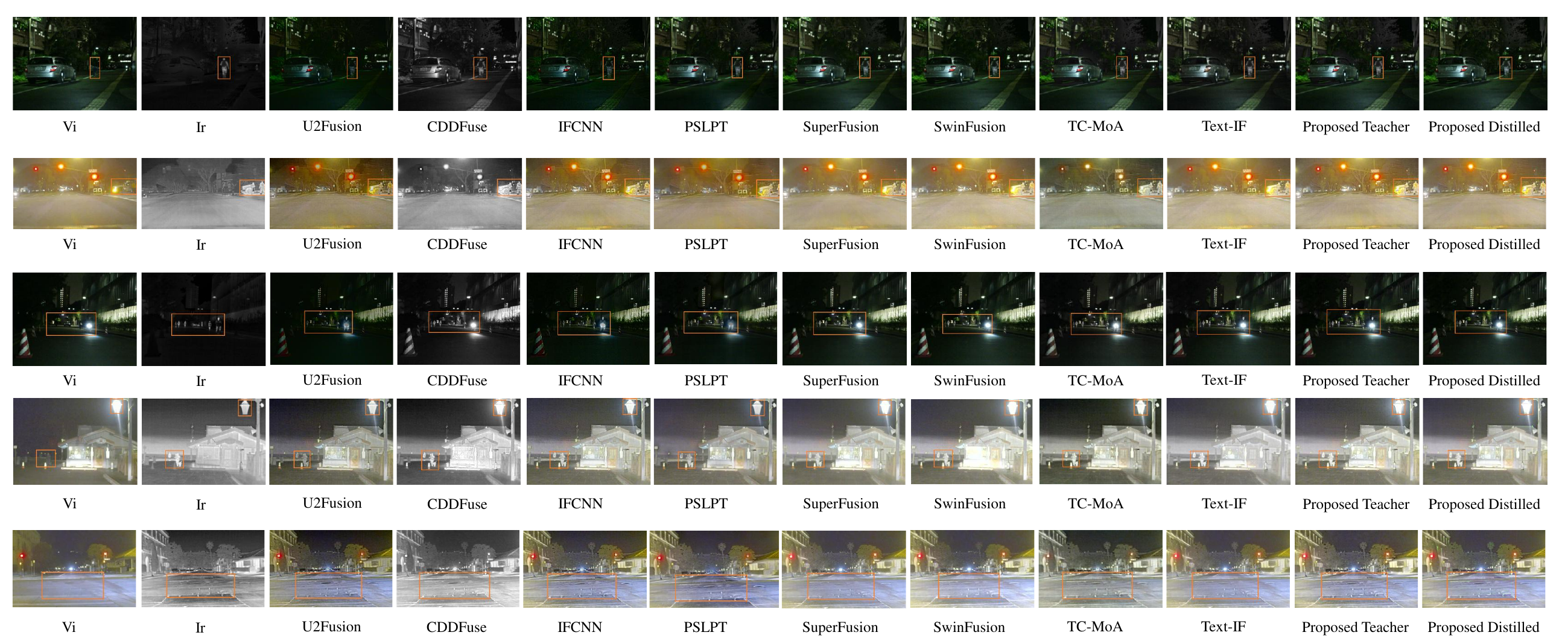}
    \caption{Comprehensive visual comparison with different methods on various IVF datasets (MSRS, M3FD, RoadScene). For each set of results, from left to right: Visible Input, Infrared Input, Text-IF Output, Ours-Teacher Output, Ours-Distilled Output.}
    \label{fig:ivf_compare_full}
\end{figure*}

\subsection{Limitations}

While our distilled model achieves significant inference-time efficiency, we acknowledge a limitation in training overhead. The knowledge distillation process introduces additional computational costs during the training phase.

\begin{table}[h]
    \centering
    \caption{Training time comparison for different epochs. All experiments were conducted on a single NVIDIA 4090 GPU.}
    \label{tab:training_time}
    \begin{tabular}{l|cc}
        \toprule
        Method & 100 epochs & 500 epochs \\
        \midrule
        Text-IF & 1.9h & 11.1h \\
        Teacher & 2.0h & 12.1h \\
        Student (distill) & 2.1h & 13.5h \\
        \bottomrule
    \end{tabular}
\end{table}

As shown in Table~\ref{tab:training_time}, our distillation approach requires slightly longer training time compared to both the baseline Text-IF and the teacher model. For a 100-epoch training cycle, the student network's distillation process takes 2.1 hours, approximately 10.5\% longer than Text-IF (1.9h) and 5\% longer than the teacher model (2.0h). This pattern persists for longer training durations, with 500-epoch training requiring 13.5 hours for distillation compared to 11.1 hours for Text-IF and 12.1 hours for the teacher model.

This increased training time is primarily attributed to two factors: (1) the two-stage training process where the teacher must be trained first, and (2) the additional computational overhead from the distillation loss calculations. However, we consider this a reasonable trade-off given the substantial inference-time benefits, as the training process is typically a one-time cost while inference efficiency directly impacts real-world applications.

\section{Conclusion}
We address the weak semantic understanding of traditional image fusion methods and the high cost of text-guided approaches by distilling textual priors and eliminating the need for text guidance during inference. Our approach, which leverages a teacher-student architecture and tailored prior distillation, significantly reduces the model size while retaining high performance. Extensive experiments and ablation studies validate the effectiveness of our method, highlighting its ability to achieve a strong trade-off between computational efficiency and fusion quality.

\nocite{*}
\bibliographystyle{IEEEtran}
\bibliography{references}

\begin{thebibliography}{10}
\providecommand{\url}[1]{#1}
\csname url@samestyle\endcsname
\providecommand{\newblock}{\relax}
\providecommand{\bibinfo}[2]{#2}
\providecommand{\BIBentrySTDinterwordspacing}{\spaceskip=0pt\relax}
\providecommand{\BIBentryALTinterwordstretchfactor}{4}
\providecommand{\BIBentryALTinterwordspacing}{\spaceskip=\fontdimen2\font plus
\BIBentryALTinterwordstretchfactor\fontdimen3\font minus \fontdimen4\font\relax}
\providecommand{\BIBforeignlanguage}[2]{{%
\expandafter\ifx\csname l@#1\endcsname\relax
\typeout{** WARNING: IEEEtran.bst: No hyphenation pattern has been}%
\typeout{** loaded for the language `#1'. Using the pattern for}%
\typeout{** the default language instead.}%
\else
\language=\csname l@#1\endcsname
\fi
#2}}
\providecommand{\BIBdecl}{\relax}
\BIBdecl

\bibitem{Zhang2025tcsvt}
J.~Zhang, K.~Cao, K.~Yan, Y.~Lin, X.~He, Y.~Wang, R.~Li, C.~Xie, J.~Zhang, and M.~Zhou, ``Frequency decoupled domain-irrelevant feature learning for pan-sharpening,'' \emph{IEEE Transactions on Circuits and Systems for Video Technology}, vol.~35, no.~2, pp. 1237--1250, 2025.

\bibitem{ma2019fusionGAN}
\BIBentryALTinterwordspacing
J.~Ma, W.~Yu, P.~Liang, C.~Li, and J.~Jiang, ``Fusiongan: A generative adversarial network for infrared and visible image fusion,'' \emph{Information Fusion}, vol.~48, pp. 11--26, 2019. [Online]. Available: \url{https://www.sciencedirect.com/science/article/pii/S1566253518301143}
\BIBentrySTDinterwordspacing

\bibitem{xu2020u2fusion}
H.~Xu, J.~Ma, J.~Jiang, X.~Guo, and H.~Ling, ``U2fusion: A unified unsupervised image fusion network,'' \emph{IEEE Transactions on Pattern Analysis and Machine Intelligence}, vol.~44, no.~1, pp. 502--518, 2020.

\bibitem{ma2022swinfusion}
J.~Ma, L.~Tang, F.~Fan, J.~Huang, X.~Mei, and Y.~Ma, ``Swinfusion: Cross-domain long-range learning for general image fusion via swin transformer,'' \emph{IEEE/CAA Journal of Automatica Sinica}, vol.~9, no.~7, pp. 1200--1217, 2022.

\bibitem{tang2022piafu}
\BIBentryALTinterwordspacing
L.~Tang, J.~Yuan, H.~Zhang, X.~Jiang, and J.~Ma, ``Piafusion: A progressive infrared and visible image fusion network based on illumination aware,'' \emph{Information Fusion}, vol. 83-84, pp. 79--92, 2022. [Online]. Available: \url{https://www.sciencedirect.com/science/article/pii/S156625352200032X}
\BIBentrySTDinterwordspacing

\bibitem{cheng2023textfusion}
C.~Cheng, T.~Xu, X.-J. Wu, H.~Li, X.~Li, Z.~Tang, and J.~Kittler, ``Textfusion: Unveiling the power of textual semantics for controllable image fusion,'' \emph{arXiv preprint arXiv:2312.14209}, 2023.

\bibitem{yi2024text}
X.~Yi, H.~Xu, H.~Zhang, L.~Tang, and J.~Ma, ``Text-if: Leveraging semantic text guidance for degradation-aware and interactive image fusion,'' in \emph{Proceedings of the IEEE/CVF Conference on Computer Vision and Pattern Recognition (CVPR)}, 2024.

\bibitem{Qwen-VL}
J.~Bai, S.~Bai, S.~Yang, S.~Wang, S.~Tan, P.~Wang, J.~Lin, C.~Zhou, and J.~Zhou, ``Qwen-vl: A versatile vision-language model for understanding, localization, text reading, and beyond,'' \emph{arXiv preprint arXiv:2308.12966}, 2023.

\bibitem{clip}
\BIBentryALTinterwordspacing
A.~Radford, J.~W. Kim, C.~Hallacy, A.~Ramesh, G.~Goh, S.~Agarwal, G.~Sastry, A.~Askell, P.~Mishkin, J.~Clark, G.~Krueger, and I.~Sutskever, ``Learning transferable visual models from natural language supervision,'' \emph{CoRR}, vol. abs/2103.00020, 2021. [Online]. Available: \url{https://arxiv.org/abs/2103.00020}
\BIBentrySTDinterwordspacing

\bibitem{zhao2023DDFM}
Z.~Zhao, H.~Bai, Y.~Zhu, J.~Zhang, S.~Xu, Y.~Zhang, K.~Zhang, D.~Meng, R.~Timofte, and L.~Van~Gool, ``Ddfm: Denoising diffusion model for multi-modality image fusion,'' in \emph{Proceedings of the IEEE/CVF International Conference on Computer Vision (ICCV)}, October 2023, pp. 8082--8093.

\bibitem{zhou2024probing}
M.~Zhou, N.~Zheng, X.~He, D.~Hong, and J.~Chanussot, ``Probing synergistic high-order interaction for multi-modal image fusion,'' \emph{IEEE Transactions on Pattern Analysis and Machine Intelligence}, 2024.

\bibitem{cao2023multi}
B.~Cao, Y.~Sun, P.~Zhu, and Q.~Hu, ``Multi-modal gated mixture of local-to-global experts for dynamic image fusion,'' in \emph{Proceedings of the IEEE/CVF International Conference on Computer Vision}, 2023, pp. 23\,555--23\,564.

\bibitem{hinton2015distilling}
G.~Hinton, ``Distilling the knowledge in a neural network,'' \emph{arXiv preprint arXiv:1503.02531}, 2015.

\bibitem{yuan2020revisiting}
L.~Yuan, F.~E. Tay, G.~Li, T.~Wang, and J.~Feng, ``Revisiting knowledge distillation via label smoothing regularization,'' in \emph{Proceedings of the IEEE/CVF conference on computer vision and pattern recognition}, 2020, pp. 3903--3911.

\bibitem{wu2023tinyclip}
K.~Wu, H.~Peng, Z.~Zhou, B.~Xiao, M.~Liu, L.~Yuan, H.~Xuan, M.~Valenzuela, X.~S. Chen, X.~Wang \emph{et~al.}, ``Tinyclip: Clip distillation via affinity mimicking and weight inheritance,'' in \emph{Proceedings of the IEEE/CVF International Conference on Computer Vision}, 2023, pp. 21\,970--21\,980.

\bibitem{zhao2023mobilediffusion}
Y.~Zhao, Y.~Xu, Z.~Xiao, and T.~Hou, ``Mobilediffusion: Subsecond text-to-image generation on mobile devices,'' \emph{arXiv preprint arXiv:2311.16567}, 2023.

\bibitem{chen2025comparative}
X.~Chen, Z.~Li, Y.~Pu, Y.~Liu, J.~Zhou, Y.~Qiao, and C.~Dong, ``A comparative study of image restoration networks for general backbone network design,'' in \emph{European Conference on Computer Vision}.\hskip 1em plus 0.5em minus 0.4em\relax Springer, 2025, pp. 74--91.

\bibitem{Zamir2021Restormer}
S.~W. Zamir, A.~Arora, S.~Khan, M.~Hayat, F.~S. Khan, and M.-H. Yang, ``Restormer: Efficient transformer for high-resolution image restoration,'' in \emph{CVPR}, 2022.

\bibitem{liu2021swin}
Z.~Liu, Y.~Lin, Y.~Cao, H.~Hu, Y.~Wei, Z.~Zhang, S.~Lin, and B.~Guo, ``Swin transformer: Hierarchical vision transformer using shifted windows,'' in \emph{Proceedings of the IEEE/CVF international conference on computer vision}, 2021, pp. 10\,012--10\,022.

\bibitem{SSIM}
Z.~Wang, A.~Bovik, H.~Sheikh, and E.~Simoncelli, ``Image quality assessment: from error visibility to structural similarity,'' \emph{IEEE Transactions on Image Processing}, vol.~13, no.~4, pp. 600--612, 2004.

\bibitem{Qwen2VL}
P.~Wang, S.~Bai, S.~Tan, S.~Wang, Z.~Fan, J.~Bai, K.~Chen, X.~Liu, J.~Wang, W.~Ge, Y.~Fan, K.~Dang, M.~Du, X.~Ren, R.~Men, D.~Liu, C.~Zhou, J.~Zhou, and J.~Lin, ``Qwen2-vl: Enhancing vision-language model's perception of the world at any resolution,'' \emph{arXiv preprint arXiv:2409.12191}, 2024.

\bibitem{MSRS_dataset}
L.~Tang, J.~Yuan, H.~Zhang, X.~Jiang, and J.~Ma, ``{MSRS: Multi-Spectral Road Scenarios for Practical Infrared and Visible Image Fusion},'' \url{https://github.com/Linfeng-Tang/MSRS}, 2022.

\bibitem{liu2022target}
J.~Liu, X.~Fan, Z.~Huang, G.~Wu, R.~Liu, W.~Zhong, and Z.~Luo, ``Target-aware dual adversarial learning and a multi-scenario multi-modality benchmark to fuse infrared and visible for object detection,'' in \emph{Proceedings of the IEEE/CVF Conference on Computer Vision and Pattern Recognition}, 2022, pp. 5802--5811.

\bibitem{roadscene}
H.~Xu, J.~Ma, Z.~Le, J.~Jiang, and X.~Guo, ``Fusiondn: A unified densely connected network for image fusion,'' in \emph{proceedings of the Thirty-Fourth AAAI Conference on Artificial Intelligence}, 2020.

\bibitem{EN}
\BIBentryALTinterwordspacing
J.~W. Roberts, J.~A. van Aardt, and F.~B. Ahmed, ``{Assessment of image fusion procedures using entropy, image quality, and multispectral classification},'' \emph{Journal of Applied Remote Sensing}, vol.~2, no.~1, p. 023522, 2008. [Online]. Available: \url{https://doi.org/10.1117/1.2945910}
\BIBentrySTDinterwordspacing

\bibitem{VIF}
\BIBentryALTinterwordspacing
Y.~Han, Y.~Cai, Y.~Cao, and X.~Xu, ``A new image fusion performance metric based on visual information fidelity,'' \emph{Information Fusion}, vol.~14, no.~2, pp. 127--135, 2013. [Online]. Available: \url{https://www.sciencedirect.com/science/article/pii/S156625351100056X}
\BIBentrySTDinterwordspacing

\bibitem{Qabf}
G.~Piella and H.~Heijmans, ``A new quality metric for image fusion,'' in \emph{Proceedings 2003 International Conference on Image Processing (Cat. No.03CH37429)}, vol.~3, 2003, pp. III--173.

\bibitem{zhao2023cddfuse}
Z.~Zhao, H.~Bai, J.~Zhang, Y.~Zhang, S.~Xu, Z.~Lin, R.~Timofte, and L.~Van~Gool, ``Cddfuse: Correlation-driven dual-branch feature decomposition for multi-modality image fusion,'' in \emph{Proceedings of the IEEE/CVF conference on computer vision and pattern recognition}, 2023, pp. 5906--5916.

\bibitem{TANG2022SuperFusion}
L.~Tang, Y.~Deng, Y.~Ma, J.~Huang, and J.~Ma, ``Superfusion: A versatile image registration and fusion network with semantic awareness,'' \emph{IEEE/CAA Journal of Automatica Sinica}, vol.~9, no.~12, pp. 2121--2137, 2022.

\bibitem{zhang2020IFCNN}
Y.~Zhang, Y.~Liu, P.~Sun, H.~Yan, X.~Zhao, and L.~Zhang, ``Ifcnn: A general image fusion framework based on convolutional neural network,'' \emph{Information Fusion}, vol.~54, pp. 99--118, 2020.

\bibitem{zhu2024tcmoa}
P.~Zhu, Y.~Sun, B.~Cao, and Q.~Hu, ``Task-customized mixture of adapters for general image fusion,'' in \emph{Proceedings of the IEEE/CVF Conference on Computer Vision and Pattern Recognition}, 2024.

\bibitem{wang2024pslpt}
W.~Wang, L.-J. Deng, and G.~Vivone, ``A general image fusion framework using multi-task semi-supervised learning,'' \emph{Information Fusion}, p. 102414, 2024.

\bibitem{xu2021emfusion}
H.~Xu and J.~Ma, ``Emfusion: An unsupervised enhanced medical image fusion network,'' \emph{Information Fusion}, 2021.

\bibitem{zero}
F.~Lahoud and S.~Süsstrunk, ``Zero-learning fast medical image fusion,'' in \emph{2019 22th International Conference on Information Fusion (FUSION)}, 2019, pp. 1--8.

\bibitem{MSRPAN}
\BIBentryALTinterwordspacing
J.~Fu, W.~Li, J.~Du, and Y.~Huang, ``A multiscale residual pyramid attention network for medical image fusion,'' \emph{Biomedical Signal Processing and Control}, vol.~66, p. 102488, 2021. [Online]. Available: \url{https://www.sciencedirect.com/science/article/pii/S1746809421000859}
\BIBentrySTDinterwordspacing

\bibitem{dao2023flashattention2}
T.~Dao, ``Flash{A}ttention-2: Faster attention with better parallelism and work partitioning,'' in \emph{International Conference on Learning Representations (ICLR)}, 2024.

\bibitem{chen2024xrestormer}
\BIBentryALTinterwordspacing
X.~Chen, Z.~Li, Y.~Pu, Y.~Liu, J.~Zhou, Y.~Qiao, and C.~Dong, ``A comparative study of image restoration networks for general backbone network design,'' 2024. [Online]. Available: \url{https://arxiv.org/abs/2310.11881}
\BIBentrySTDinterwordspacing

\bibitem{op2021styleclip}
\BIBentryALTinterwordspacing
O.~Patashnik, Z.~Wu, E.~Shechtman, D.~Cohen{-}Or, and D.~Lischinski, ``Styleclip: Text-driven manipulation of stylegan imagery,'' \emph{CoRR}, vol. abs/2103.17249, 2021. [Online]. Available: \url{https://arxiv.org/abs/2103.17249}
\BIBentrySTDinterwordspacing

\bibitem{rombach2021highresolution}
R.~Rombach, A.~Blattmann, D.~Lorenz, P.~Esser, and B.~Ommer, ``High-resolution image synthesis with latent diffusion models,'' 2021.

\bibitem{SF}
V.~Petrovic and C.~Xydeas, ``Objective image fusion performance characterisation,'' in \emph{Tenth IEEE International Conference on Computer Vision (ICCV'05) Volume 1}, vol.~2, 2005, pp. 1866--1871 Vol. 2.

\bibitem{SD}
X.~X. Zhu and R.~Bamler, ``A sparse image fusion algorithm with application to pan-sharpening,'' \emph{IEEE Transactions on Geoscience and Remote Sensing}, vol.~51, no.~5, pp. 2827--2836, 2013.

\bibitem{guo2024benchmarking}
D.~Guo, K.~Li, B.~Hu, Y.~Zhang, and M.~Wang, ``Benchmarking micro-action recognition: Dataset, methods, and applications,'' \emph{IEEE Transactions on Circuits and Systems for Video Technology}, vol.~34, no.~7, pp. 6238--6252, 2024.

\bibitem{Albarqaan2024Image}
\BIBentryALTinterwordspacing
H.~Albarqaan, R.~Qin, and Y.~Teng, ``{Image Fusion in Remote Sensing: An Overview and Meta-Analysis},'' \emph{Photogrammetric Engineering \& Remote Sensing}, vol.~90, no.~12, pp. 755--775, December 2024. [Online]. Available: \url{https://doi.org/10.14358/PERS.24-00110R1}
\BIBentrySTDinterwordspacing

\bibitem{MI}
\BIBentryALTinterwordspacing
J.~Ma, Y.~Ma, and C.~Li, ``Infrared and visible image fusion methods and applications: A survey,'' \emph{Information Fusion}, vol.~45, pp. 153--178, 2019. [Online]. Available: \url{https://www.sciencedirect.com/science/article/pii/S1566253517307972}
\BIBentrySTDinterwordspacing

\bibitem{dao2022flashattention}
T.~Dao, D.~Y. Fu, S.~Ermon, A.~Rudra, and C.~R{\'e}, ``Flash{A}ttention: Fast and memory-efficient exact attention with {IO}-awareness,'' in \emph{Advances in Neural Information Processing Systems (NeurIPS)}, 2022.

\bibitem{Zhao2024tufusion}
Y.~Zhao, Q.~Zheng, P.~Zhu, X.~Zhang, and W.~Ma, ``Tufusion: A transformer-based universal fusion algorithm for multimodal images,'' \emph{IEEE Transactions on Circuits and Systems for Video Technology}, vol.~34, no.~3, pp. 1712--1725, 2024.

\bibitem{Li2024SSSB}
X.~Li, G.~Zhang, W.~Chen, L.~Cheng, Y.~Xie, and J.~Ma, ``An infrared and visible image fusion method based on semantic-sensitive mask selection and bidirectional-collaboration region fusion,'' \emph{IEEE Transactions on Circuits and Systems for Video Technology}, pp. 1--1, 2024.

\bibitem{Zhao2019SCNN}
W.~Zhao, D.~Wang, and H.~Lu, ``Multi-focus image fusion with a natural enhancement via a joint multi-level deeply supervised convolutional neural network,'' \emph{IEEE Transactions on Circuits and Systems for Video Technology}, vol.~29, no.~4, pp. 1102--1115, 2019.

\end{thebibliography}

\end{document}